\documentclass[11pt]{article}
\usepackage{listings}

\usepackage[final]{acl}

\usepackage{times}
\usepackage{latexsym}
 \usepackage{amsmath}
\usepackage[T1]{fontenc}

\usepackage[utf8]{inputenc}

\usepackage{microtype}

\usepackage{inconsolata}

\usepackage{graphicx}

\usepackage{multirow}

\title{Tournament on Path: Global Relation Abstraction and Pair-wise Ranking for Knowledge Graph Reasoning}

\author{Yihan Wu \\
  \texttt{yihanw@chalmers.se} \\
  \And
  Shihao Xiang \\
  \texttt{shihaox@chalmers.se} \\
  \And
  Yinan Yu \\
  \texttt{yinan@chalmers.se} \\
  \And
  Shuai Wang \\  \texttt{shuaiwa@chalmers.se} \\
  }

\begin{document}
\maketitle

\begin{abstract}
While Large Language Models (LLMs) integrated with Knowledge Graphs (KGs) show promise in multi-hop reasoning, existing step-wise frameworks suffer from shortsighted pruning decisions and unstable individual candidate scoring. To address these limitations, we propose Tournament on Path (ToP), which abstracts graphs into entity-agnostic relation paths, enabling global reasoning, and introduces a tournament pair-wise ranking strategy for more robust candidate selection. Evaluated on four benchmarks, ToP consistently outperforms strong baseline and improves scaling behavior across different foundation models. Source code: \url{https://github.com/yihanwu1024/top}.
\end{abstract}

\section{Introduction}

Recent advances in Large Language Models (LLMs) have demonstrated remarkable performance across various tasks. However, they still suffer from hallucinations and factual inaccuracies in knowledge-intensive scenarios \citep{Ji_2023,huang-chang-2023-towards}. Retrieval-Augmented Generation (RAG) mitigates this issue by incorporating external knowledge during inference \citep{gao2023retrieval}. Among different knowledge sources, Knowledge Graphs (KGs) provide structured and reliable factual information, making them particularly suitable for reasoning-intensive tasks. Consequently, Knowledge Graph Question Answering (KGQA), which answers questions through reasoning over KGs, has become an important research topic \citep{ma2025large}.

For complex questions, single-step reasoning is often insufficient because the starting entity (in the question) and answer entities are usually far away in the KGs, requiring multi-hop reasoning. To address this challenge, recent methods leverage LLMs as agents to perform step-wise reasoning over KGs by iteratively exploring relations \cite{10387715}. A representative approach is Think-on-Graph (ToG)~\cite{sun2024thinkongraph}, which utilizes the reasoning capability of LLMs to guide graph traversal. At each step, ToG scores one-hop neighboring entities individually and selects the most relevant candidates for further exploration. Such methods can directly leverage off-the-shelf LLMs and treat KG reasoning as a plug-and-play component, making them simple and practical to deploy.

However, this paradigm suffers from several inherent limitations. First, relation pruning decisions are locally scoped, failing to capture the global semantic intent of complex queries. Second, pruning is implemented by point-wise scoring, introducing instability in candidate selection. Finally, ToG also resorts to random sampling to resolve the large number of neighbors in KGs, inevitably discarding useful neighbors.

To address these issues, we propose Tournament on Path (ToP) with two main improvements. First, we merge exploration paths by relation path equivalence, ignoring intermediate entities. This structure enables global path-level evaluation and reduces entity-level noise. Second, to further optimize pruning stability, we introduce a tournament strategy. This replaces unstable point-wise scoring with a pair-wise ranked tournament algorithm after a semantic pre-filter. Our main contributions are as follows:
\begin{itemize}
    \item We mitigate shortsighted pruning decisions by abstracting graphs into entity-agnostic relation paths, enabling global path-level reasoning.
    \item We design a tournament selection strategy that replaces point-wise scoring with pair-wise ranking, improving selection stability.
    \item Experiments on four benchmark datasets show that our method consistently improves reasoning robustness and pruning stability.
\end{itemize}

\section{Method}

\begin{figure}[t]

  \includegraphics[width=\columnwidth]{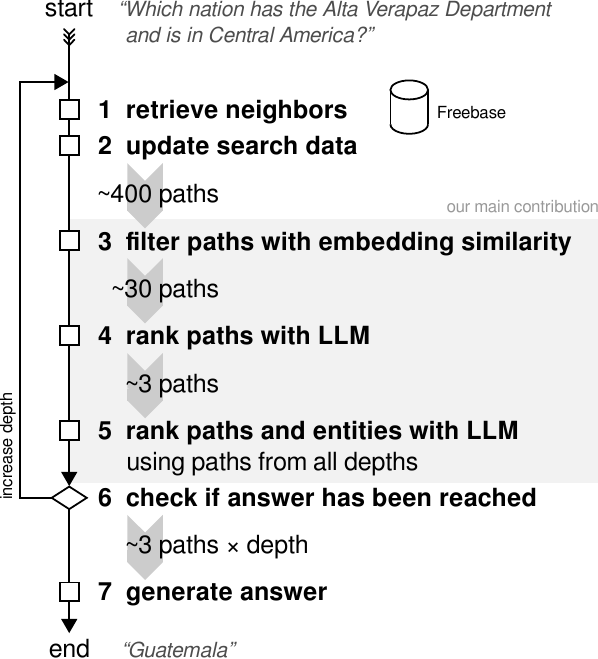}
  \caption{Overall architecture of ToP}
  \label{fig:experiments}
\end{figure}

In this section, we propose two methods within ToP, each addressing a shortcoming in the ToG architecture.
Figure \ref{fig:experiments} illustrates the overall architecture of our ToP framework.


\subsection{Entity-Agnostic Relation Path Pruning}

We observe that in multi-hop reasoning over knowledge graphs, the semantic utility of a reasoning path is largely governed by its sequence of relations. For example, \textit{Chicago $\xrightarrow{\text{located\_in}}$ USA $\xrightarrow{\text{national\_bird}}$ Bald Eagle } can be abstracted as the compositional relation $R(X,Z)$ defined over relation paths.
\begin{equation}
\begin{aligned}
R(X&, Z)  =  \\
& \exists Y, \text{located\_in}(X, Y) \land \text{national\_bird}(Y, Z)
\end{aligned}
\end{equation}

As ToG performs relation pruning step by step, it remains limited to local context and may fail to capture global path semantics.
To address this, ToP performs relation pruning using---and only using---relation information. Intermediate entities are deliberately ignored in the relation path pruning process, thereby merging full entity-aware paths by identical start entity and sequence of relations (Figure \ref{fig:relpath}). More formally, full paths of the form
$$e_0, r_1, e_1, r_2, e_2, \cdots, r_n, e_n$$
where $e$ denotes entity, $e_0$ is a given topic entity, and $r$ denotes relation,
are path-equivalent if their subsequences $e_0, r_1, r_2, \cdots, r_n$ are identical.
This approach enables more effective pruning by allowing the LLM to evaluate global path semantics while reducing entity-level noise.

This approach does have some limitations. For example, consider the question ``What is the nationality of the tallest person in the world?'' on a knowledge graph without the superlative relation ``tallest person in''. To answer this question, it is necessary to instantiate the entity referred to by ``the tallest person in the world'', which is impossible. Despite this limitation in expressiveness, we propose our method as it brings clear performance improvements and may still be extended to support this kind of expressiveness.

\subsection{Tournament Candidate Selection}

In order to improve the  stability of scoring, ToP introduces a tournament algorithm after a pre-filter.

\paragraph{Semantic Pre-filter}
On Freebase, the size of the candidate pool usually reach hundreds after exploration at each depth. The original ToG framework manages this by randomly discarding candidates to fit a fixed budget prior to LLM scoring. This inevitably discards relevant entities.

To reduce the search space while preserving relevance, ToP employs a semantic pre-filter by selecting top $m$ paths with the highest embedding similarity against the natural language question\footnote{We used $m = 30$.}.

\paragraph{LLM as Tournament Chooser, Not Scorer}

We reformulate candidate selection as an LLM-bound pair-wise selection process.

The original ToG relies on assigning absolute numerical scores, leading to scoring failures such as degrading to extreme scores ($0$ or $1$) and assigning the same score to vastly different candidates. This may be due to the absence of calibration in LLM scoring. Recent work on LLM-based evaluation has shown that relative comparative ranking improves evaluation consistency over absolute scoring \citep{zheng2026isolatedscoringcollaborativeranking, sun-etal-2023-chatgpt, lee-etal-2025-shifting}. Jointly evaluating candidates provides richer mutual context, enabling the LLM to capture fine-grained differences for improved decision-making. Building on this observation, ToP treats the LLM as a comparative chooser rather than a point-wise scorer.

\paragraph{LLM-Guided Tournament Selection}

Our tournament algorithm is a version of the well-documented merge sort algorithm, where the LLM powers the compare function. For our purpose, we do not require LLM comparison to be consistent (for example, $a > b > c > a$ is permissible) or transitive, as our goal is to exploit the LLM's reasoning capability enough to probabilistically find the more accurate paths.

Traditionally, the merge step in merge sort creates an empty new list $l$, compares the top item from each of the two input lists, moves the greater one to the new list $l$, and iterates until one input list is empty, in which case all remaining items are directly added to the new list $l$.
We modify merge sort to limit the item count in $l$ to some $k$ during this merge step\footnote{We used $k = 3$.}.
The modified merge routine ends once the merged list $l$ reaches size $k$, saving time that would have been used to compare remaining (worse) candidates.
This algorithm has a time complexity of $O(n \log n)$ if all comparisons are serial.

\subsection{Prompt Design}

Our prompt design reflects the architectural shifts introduced above: ignoring entities in relation paths for prompting, and replacing point-wise scoring with pair-wise selection.
The design of our prompts can be found in the appendix \ref{sec:prompt}.

\paragraph{Diff prompting for path comparison} In order to trick the LLM into comparison, we symbolically extract the longest common prefix of the two path candidates. This common prefix is presented as ``given conditions'' to the LLM. The remaining segment of each path, the \textit{path diff}, each becomes an option. In case of any empty path diff, the question is formulated as a relevance decision problem and only the nonempty path diff is presented.

\section{Experiment}

\begin{table*}
  \centering
  \begin{tabular}{llcccc}
    \hline
    \textbf{Models}    &       & \textbf{CWQ} & \textbf{WebQ} & \textbf{WebQSP} & \textbf{GrailQA} \\
    \hline
    \multirow{ 2}{*}{Qwen 3 32B} & w/ ToG & 38.0 & 46.0 & 59.5 & 50.0 \\ 
              & w/ ToP & 43.5 & 63.5 & 73.0 & 74.0 \\
    \hline
    \multirow{ 2}{*}{Qwen 3 14B} & w/ ToG & 34.0 & 54.0 & 62.5 & 56.0 \\ 
              & w/ ToP & 37.5 & 57.0 & 62.5 & 59.5 \\
    \hline
    \multirow{ 2}{*}{Gemma 4 E4B IT} & w/ ToG & 31.0 & 28.0 & 39.0 & 53.5 \\ 
              & w/ ToP & 46.0 & 56.5 & 67.5 & 74.0 \\
    \hline
    \multirow{ 2}{*}{Gemma 4 E2B IT} & w/ ToG & 12.0 & 17.0 & 22.0 & 10.5 \\ 
              & w/ ToP & 39.5 & 52.0 & 64.5 & 67.5 \\
    \hline
  \end{tabular}
  \caption{\label{result}
     Reasoning performance of various LLMs equipped with ToP versus ToG. The results demonstrate that ToP not only achieves an overall improvement over ToG, but also scales with LLM size and exhibits consistency across datasets. }
\end{table*}

In this section, we focus on the following research questions.

\begin{enumerate}
    \item To what extent does the proposed framework improve multi-hop reasoning performance over standard step-wise baseline?
    \item How does model scale affect reasoning performance, and how does the proposed method reshape this scaling behavior?
    \item How consistently does the proposed framework perform across datasets with varying schema complexities, compared to the baseline?
\end{enumerate}

To answer these questions, we first detail our experiment setup, and then conduct a detailed analysis across three dimensions.

\subsection{Setup}

\paragraph{Datasets and Metrics} Following \cite{sun2024thinkongraph}, we employ Freebase \cite{10.1145/1376616.1376746} as our underlying knowledge graph. We evaluate our method across four widely recognized benchmarks that feature reasoning tasks of different complexities: CWQ \cite{talmor-berant-2018-web}, WebQuestions \cite{berant-etal-2013-semantic} (WebQ), WebQSP \cite{yih-etal-2016-value},  and GrailQA \cite{10.1145/3442381.3449992}. To optimize computational efficiency while maintaining evaluation integrity, we randomly sample a subset of 200 questions from each dataset for our evaluation. We adopt \textbf{Exact Match} (EM) as our primary metric to determine whether the predicted answer aligns with the ground-truth annotations.

\paragraph{Language Models} To demonstrate the generalization and robustness of our proposed method across different scales of LLMs, we conduct experiments using four representative small to medium-sized open-source reasoning models: Qwen 3 14B, 32B \cite{qwen3technicalreport}; and Gemma 4 E2B IT, E4B IT (instruction-tuned) \cite{gemma4modelcard}. The pre-filter in ToP is Qwen 3 Embedding 4B \cite{qwen3embedding} for all experiments.

\subsection{Analysis}

We conduct an analysis across three distinct dimensions. The results are shown in Table \ref{result}.

\subsubsection{Performance Gains Across Frameworks}

Our experiment results show that ToP consistently outperforms the standard ToG baseline across nearly all tested LLM and benchmarks. Notably, when paired with smaller models, such as Gemma 4 E2B, ToP achieves substantial EM improvements of 27.5\% on CWQ, 35.0\% on WebQ, 42.5\% on WebQSP and 57.0\% on GrailQA. Even for larger models like Qwen 3 32B, ToP still yields significant gains, improving performance by up to 24.0\% absolute on GrailQA.

These empirical improvements suggest that restructuring step-wise path pruning into path-level reasoning in conjunction with tournament candidate selection leads to more effective filtering of reasoning paths, especially for small LLMs.

\subsubsection{Impact of LLM Model Scale and Capability}

We observed that the baseline ToG results show a certain degree of performance inversion, or ``negative scaling'', among Qwen models as the parameter size increases. For instance, the larger Qwen 3 32B underperforms the smaller Qwen 3 14B across multiple datasets, including WebQ (46.0\% vs 54.0\%), GrailQA (50.0\% vs 56.0\%), and WebQSP (59.5\% vs 62.5\%). We attribute this to longer and more flexible responses, which more frequently violate formatting requirements, generated by larger models.

There is an apparent performance inversion between the Qwen 3 and Gemma 4 LLM families, specific to ToP and not ToG. We attribute this to the different sensitivities of LLMs to certain prompt styles. During pruning, ToG asks scoring questions with a list of candidates using few-shot prompting, while ToP asks zero-shot multiple-choice questions with exactly two options and requires the LLM to show its reasoning process. The choice of an instruction-tuned model (Gemma 4 E4B IT and E2B IT) may also have made a difference. Finally, it is worth noting that the Gemma 4 LLM family is approximately one year more recent than Qwen 3, entailing non-trivial systematic changes.

\subsubsection{Performance Consistency Across Datasets}


ToP exhibits stable performance across different benchmarks, with variations aligned with the baseline.
This pattern indicates that ToP provides consistent behavior across datasets rather than being optimized to a specific benchmark, and the variations should be explained by differences in question complexity.

\subsection{Ablation Study}

\begin{table}[t!]
  \centering
  \begin{tabular}{lcccc}
    \hline
    \textbf{Methods}          & \textbf{CWQ} & \textbf{WebQ} \\
    \hline
     ToP & 46.0 & 56.5 \\
     ToP w/o Relation Path & 33.0 & 45.5   \\
     ToP w/o Tournament & 32.5 & 53.0  \\ 

    \hline
  \end{tabular}
  \caption{\label{ablation-table}
    Ablation study under Gemma 4 E4B IT.
  }
\end{table}

Table \ref{ablation-table} presents the ablation study results of the ToP framework using Gemma 4 E4B IT to evaluate our two core components.

\paragraph{Ablation of relation path pruning} Only the final relation in the relation path is used for pruning.

\paragraph{Ablation of tournament} The candidates are point-wise scored rather than ranked by pair-wise comparison. We batch 5 items per LLM prompt. Scoring is implemented for both entity-agnostic path pruning and final path ranking with entity information.

\paragraph{} Both ablations reduce performance from full ToP on CWQ and WebQ, demonstrating the necessity of these modifications.

\section{Conclusion}

In this work, we introduced Tournament on Path, a novel framework that improves upon current paradigms through two key mechanisms: abstracting exploration paths into entity-agnostic relation sequences for global path-level reasoning, and replacing unstable point-wise scoring with a robust pair-wise tournament selection algorithm. Extensive experiments across four KGQA benchmarks and four LLMs demonstrate that ToP consistently outperforms baseline methods and scales positively across different models. 

 \newpage

\section*{Limitations}

Although our approach mitigates shortsighted pruning decisions by taking advantage of a global evaluation perspective, the reasoning process remains grounded in the connectivity of the graph. Consequently, if critical relation paths or entities are missing from the underlying KG, the structural search space becomes fragmented. In such scenarios, even ToP cannot retrieve the necessary evidence to complete the reasoning chain, compromising the overall performance.



\bibliography{custom}

@inproceedings{ma2025large,
    title = "Large Language Models Meet Knowledge Graphs for Question Answering: Synthesis and Opportunities",
    author = "Ma, Chuangtao  and
      Chen, Yongrui  and
      Wu, Tianxing  and
      Khan, Arijit  and
      Wang, Haofen",
    editor = "Christodoulopoulos, Christos  and
      Chakraborty, Tanmoy  and
      Rose, Carolyn  and
      Peng, Violet",
    booktitle = "Proceedings of the 2025 Conference on Empirical Methods in Natural Language Processing",
    month = nov,
    year = "2025",
    address = "Suzhou, China",
    publisher = "Association for Computational Linguistics",
    url = "https://aclanthology.org/2025.emnlp-main.1249/",
    doi = "10.18653/v1/2025.emnlp-main.1249",
    pages = "24578--24597",
    ISBN = "979-8-89176-332-6"
}

@misc{gao2023retrieval,
      title={Retrieval-Augmented Generation for Large Language Models: A Survey}, 
      author={Yunfan Gao and Yun Xiong and Xinyu Gao and Kangxiang Jia and Jinliu Pan and Yuxi Bi and Yi Dai and Jiawei Sun and Meng Wang and Haofen Wang},
      year={2024},
      eprint={2312.10997},
      archivePrefix={arXiv},
      primaryClass={cs.CL},
      url={https://arxiv.org/abs/2312.10997}, 
}

@misc{wang2026kgreasonerreinforcedmodelendtoend,
      title={KG-Reasoner: A Reinforced Model for End-to-End Multi-Hop Knowledge Graph Reasoning}, 
      author={Shuai Wang and Yinan Yu},
      year={2026},
      eprint={2604.12487},
      archivePrefix={arXiv},
      primaryClass={cs.CL},
      url={https://arxiv.org/abs/2604.12487}, 
}

@inproceedings{lee-etal-2025-shifting,
    title = "Shifting from Ranking to Set Selection for Retrieval Augmented Generation",
    author = "Lee, Dahyun  and
      Jo, Yongrae  and
      Park, Haeju  and
      Lee, Moontae",
    editor = "Che, Wanxiang  and
      Nabende, Joyce  and
      Shutova, Ekaterina  and
      Pilehvar, Mohammad Taher",
    booktitle = "Proceedings of the 63rd Annual Meeting of the Association for Computational Linguistics (Volume 1: Long Papers)",
    month = jul,
    year = "2025",
    address = "Vienna, Austria",
    publisher = "Association for Computational Linguistics",
    url = "https://aclanthology.org/2025.acl-long.861/",
    doi = "10.18653/v1/2025.acl-long.861",
    pages = "17606--17619",
    ISBN = "979-8-89176-251-0"
}

@misc{qwen3technicalreport,
      title={Qwen3 Technical Report}, 
      author={An Yang and Anfeng Li and Baosong Yang and Beichen Zhang and Binyuan Hui and Bo Zheng and Bowen Yu and Chang Gao and Chengen Huang and Chenxu Lv and Chujie Zheng and Dayiheng Liu and Fan Zhou and Fei Huang and Feng Hu and Hao Ge and Haoran Wei and Huan Lin and Jialong Tang and Jian Yang and Jianhong Tu and Jianwei Zhang and Jianxin Yang and Jiaxi Yang and Jing Zhou and Jingren Zhou and Junyang Lin and Kai Dang and Keqin Bao and Kexin Yang and Le Yu and Lianghao Deng and Mei Li and Mingfeng Xue and Mingze Li and Pei Zhang and Peng Wang and Qin Zhu and Rui Men and Ruize Gao and Shixuan Liu and Shuang Luo and Tianhao Li and Tianyi Tang and Wenbiao Yin and Xingzhang Ren and Xinyu Wang and Xinyu Zhang and Xuancheng Ren and Yang Fan and Yang Su and Yichang Zhang and Yinger Zhang and Yu Wan and Yuqiong Liu and Zekun Wang and Zeyu Cui and Zhenru Zhang and Zhipeng Zhou and Zihan Qiu},
      year={2025},
      eprint={2505.09388},
      archivePrefix={arXiv},
      primaryClass={cs.CL},
      url={https://arxiv.org/abs/2505.09388}, 
}

@ARTICLE{10387715,
author={Pan, Shirui and Luo, Linhao and Wang, Yufei and Chen, Chen and Wang, Jiapu and Wu, Xindong},
journal={ IEEE Transactions on Knowledge \& Data Engineering },
title={{ Unifying Large Language Models and Knowledge Graphs: A Roadmap }},
year={2024},
volume={36},
number={07},
ISSN={1558-2191},
pages={3580-3599},
doi={10.1109/TKDE.2024.3352100},
url = {https://doi.ieeecomputersociety.org/10.1109/TKDE.2024.3352100},
publisher={IEEE Computer Society},
address={Los Alamitos, CA, USA},
month=jul}

@inproceedings{sun-etal-2023-chatgpt,
    title = "Is {C}hat{GPT} Good at Search? Investigating Large Language Models as Re-Ranking Agents",
    author = "Sun, Weiwei  and
      Yan, Lingyong  and
      Ma, Xinyu  and
      Wang, Shuaiqiang  and
      Ren, Pengjie  and
      Chen, Zhumin  and
      Yin, Dawei  and
      Ren, Zhaochun",
    editor = "Bouamor, Houda  and
      Pino, Juan  and
      Bali, Kalika",
    booktitle = "Proceedings of the 2023 Conference on Empirical Methods in Natural Language Processing",
    month = dec,
    year = "2023",
    address = "Singapore",
    publisher = "Association for Computational Linguistics",
    url = "https://aclanthology.org/2023.emnlp-main.923/",
    doi = "10.18653/v1/2023.emnlp-main.923",
    pages = "14918--14937"
}

@inproceedings{10.1145/3442381.3449992,
author = {Gu, Yu and Kase, Sue and Vanni, Michelle and Sadler, Brian and Liang, Percy and Yan, Xifeng and Su, Yu},
title = {Beyond I.I.D.: Three Levels of Generalization for Question Answering on Knowledge Bases},
year = {2021},
isbn = {9781450383127},
publisher = {Association for Computing Machinery},
address = {New York, NY, USA},
url = {https://doi.org/10.1145/3442381.3449992},
doi = {10.1145/3442381.3449992},
booktitle = {Proceedings of the Web Conference 2021},
pages = {3477–3488},
numpages = {12},
location = {Ljubljana, Slovenia},
series = {WWW '21}
}

@inproceedings{berant-etal-2013-semantic,
    title = "Semantic Parsing on {F}reebase from Question-Answer Pairs",
    author = "Berant, Jonathan  and
      Chou, Andrew  and
      Frostig, Roy  and
      Liang, Percy",
    editor = "Yarowsky, David  and
      Baldwin, Timothy  and
      Korhonen, Anna  and
      Livescu, Karen  and
      Bethard, Steven",
    booktitle = "Proceedings of the 2013 Conference on Empirical Methods in Natural Language Processing",
    month = oct,
    year = "2013",
    address = "Seattle, Washington, USA",
    publisher = "Association for Computational Linguistics",
    url = "https://aclanthology.org/D13-1160/",
    pages = "1533--1544"
}

@inproceedings{yih-etal-2016-value,
    title = "The Value of Semantic Parse Labeling for Knowledge Base Question Answering",
    author = "Yih, Wen-tau  and
      Richardson, Matthew  and
      Meek, Chris  and
      Chang, Ming-Wei  and
      Suh, Jina",
    editor = "Erk, Katrin  and
      Smith, Noah A.",
    booktitle = "Proceedings of the 54th Annual Meeting of the Association for Computational Linguistics (Volume 2: Short Papers)",
    month = aug,
    year = "2016",
    address = "Berlin, Germany",
    publisher = "Association for Computational Linguistics",
    url = "https://aclanthology.org/P16-2033/",
    doi = "10.18653/v1/P16-2033",
    pages = "201--206"
}

@inproceedings{talmor-berant-2018-web,
    title = "The Web as a Knowledge-Base for Answering Complex Questions",
    author = "Talmor, Alon  and
      Berant, Jonathan",
    editor = "Walker, Marilyn  and
      Ji, Heng  and
      Stent, Amanda",
    booktitle = "Proceedings of the 2018 Conference of the North {A}merican Chapter of the Association for Computational Linguistics: Human Language Technologies, Volume 1 (Long Papers)",
    month = jun,
    year = "2018",
    address = "New Orleans, Louisiana",
    publisher = "Association for Computational Linguistics",
    url = "https://aclanthology.org/N18-1059/",
    doi = "10.18653/v1/N18-1059",
    pages = "641--651"
}

@inproceedings{10.1145/1376616.1376746,
author = {Bollacker, Kurt and Evans, Colin and Paritosh, Praveen and Sturge, Tim and Taylor, Jamie},
title = {Freebase: a collaboratively created graph database for structuring human knowledge},
year = {2008},
isbn = {9781605581026},
publisher = {Association for Computing Machinery},
address = {New York, NY, USA},
url = {https://doi.org/10.1145/1376616.1376746},
doi = {10.1145/1376616.1376746},
booktitle = {Proceedings of the 2008 ACM SIGMOD International Conference on Management of Data},
pages = {1247–1250},
numpages = {4},
location = {Vancouver, Canada},
series = {SIGMOD '08}
}

@inproceedings{
sun2024thinkongraph,
title={Think-on-Graph: Deep and Responsible Reasoning of Large Language Model on Knowledge Graph},
author={Jiashuo Sun and Chengjin Xu and Lumingyuan Tang and Saizhuo Wang and Chen Lin and Yeyun Gong and Lionel Ni and Heung-Yeung Shum and Jian Guo},
booktitle={The Twelfth International Conference on Learning Representations},
year={2024},
url={https://openreview.net/forum?id=nnVO1PvbTv}
}

@misc{zheng2026isolatedscoringcollaborativeranking,
      title={From Isolated Scoring to Collaborative Ranking: A Comparison-Native Framework for LLM-Based Paper Evaluation}, 
      author={Pujun Zheng and Jiacheng Yao and Jinquan Zheng and Chenyang Gu and Guoxiu He and Jiawei Liu and Yong Huang and Tianrui Guo and Wei Lu},
      year={2026},
      eprint={2603.17588},
      archivePrefix={arXiv},
      primaryClass={cs.IR},
      url={https://arxiv.org/abs/2603.17588}, 
}

@article{Ji_2023,
   title={Survey of Hallucination in Natural Language Generation},
   volume={55},
   ISSN={1557-7341},
   url={http://dx.doi.org/10.1145/3571730},
   DOI={10.1145/3571730},
   number={12},
   journal={ACM Computing Surveys},
   publisher={Association for Computing Machinery (ACM)},
   author={Ji, Ziwei and Lee, Nayeon and Frieske, Rita and Yu, Tiezheng and Su, Dan and Xu, Yan and Ishii, Etsuko and Bang, Ye Jin and Madotto, Andrea and Fung, Pascale},
   year={2023},
   month=mar, pages={1–38} }

@inproceedings{huang-chang-2023-towards,
    title = "Towards Reasoning in Large Language Models: A Survey",
    author = "Huang, Jie  and
      Chang, Kevin Chen-Chuan",
    editor = "Rogers, Anna  and
      Boyd-Graber, Jordan  and
      Okazaki, Naoaki",
    booktitle = "Findings of the Association for Computational Linguistics: ACL 2023",
    month = jul,
    year = "2023",
    address = "Toronto, Canada",
    publisher = "Association for Computational Linguistics",
    url = "https://aclanthology.org/2023.findings-acl.67/",
    doi = "10.18653/v1/2023.findings-acl.67",
    pages = "1049--1065"
}

@misc{gemma4modelcard,
  author       = {{Google DeepMind}},
  title        = {Gemma 4 model card},
  howpublished = {\url{https://ai.google.dev/gemma/docs/core/model_card_4}},
  year         = {2026},
  note         = {Accessed: 2026-05-25}
}

@misc{qwen3embedding,
      title={Qwen3 Embedding: Advancing Text Embedding and Reranking Through Foundation Models}, 
      author={Yanzhao Zhang and Mingxin Li and Dingkun Long and Xin Zhang and Huan Lin and Baosong Yang and Pengjun Xie and An Yang and Dayiheng Liu and Junyang Lin and Fei Huang and Jingren Zhou},
      year={2025},
      eprint={2506.05176},
      archivePrefix={arXiv},
      primaryClass={cs.CL},
      url={https://arxiv.org/abs/2506.05176}, 
}

\appendix

\section{Prompt Design}
\label{sec:prompt}


\subsection{Relation Pruning} 

As part of merge sort, our prompt asks the LLM to compare the relevance of two relation paths.

Every pair of paths has a unique longest common prefix, which is extracted and presented as ``given conditions'' to the LLM. The remaining segment of each path, the \textit{path diff}, is presented as an option.

In case of two nonempty path diffs, the following zero-shot prompt structure is used:

\begin{lstlisting}[basicstyle=\small\ttfamily, breaklines=true]
You are helping a computer program perform question answering on a knowledge graph.
Do not pick information that is redundant.
In combination with the known conditions, which one of the candidates is more relevant to the following question?
Question: m.03_dwn is the mascot for the team that last won the world series when?
Given conditions: -> GIVEN_TOPIC_ENTITY: m.03_dwn
Candidate A: -> sports.mascot.team <- common.webpage.topic
Candidate B: <- sports.sports_team.team_mascot <- award.award_nomination.award_nominee
First explain your reasons in one sentence for each option, then write the response as [A] or [B].
\end{lstlisting}

Sample response:

\begin{lstlisting}[basicstyle=\small\ttfamily, breaklines=true]
Candidate A describes a general relationship between a mascot and a webpage topic, which is too broad. Candidate B connects the team mascot to sports teams and awards, which is relevant to winning a world series.
[B]
\end{lstlisting}

In case of any empty path diff, the following zero-shot prompt structure is used:

\begin{lstlisting}[basicstyle=\small\ttfamily, breaklines=true]
You are helping a computer program perform question answering on a knowledge graph.
Do not pick information that is redundant.
In combination with the known conditions, is the candidate relevant to the following question?
Question: m.03_dwn is the mascot for the team that last won the world series when?
Given conditions: -> GIVEN_TOPIC_ENTITY: m.03_dwn <- sports.sports_team.team_mascot
Candidate: <- baseball.baseball_league.teams
First explain your reasons in one sentence for each option, then write the response as [RELEVANT] or [NOT RELEVANT].
\end{lstlisting}

Sample response:

\begin{lstlisting}[basicstyle=\small\ttfamily, breaklines=true]
The candidate is too broad and does not specifically mention the team that won the world series or the mascot mentioned in the question.
[NOT RELEVANT]
\end{lstlisting}

\subsection{Answer Generation}

For final answer generation, we present the LLM with the top-ranked paths along with the query results (entity names or values), then ask for the answer if one exists, or a ``no answer'' symbol in a zero-shot manner.

\begin{lstlisting}[basicstyle=\small\ttfamily, breaklines=true]
You are helping a computer program perform question answering on a knowledge graph.
You have access to some relation paths and their query results, and your task is to identify the answer, or report if it does not exist here.
If some paths provide the answer, simply give the answer.
If no path provides any answer, write [NO ANSWER].

Question:
m.03_dwn is the mascot for the team that last won the world series when?

Paths:
-> GIVEN_TOPIC_ENTITY: m.03_dwn <- sports.sports_team.team_mascot <- sports.sports_championship_event.champion query result: [['2010 World Series'], ['2012 World Series'], ['2014 World Series']] and 0 unnamed objects
-> GIVEN_TOPIC_ENTITY: m.03_dwn <- sports.sports_team.team_mascot <- sports.sports_league_participation.team query result: [] and 3 unnamed objects
-> GIVEN_TOPIC_ENTITY: m.03_dwn <- sports.sports_team.team_mascot query result: [['San Francisco Giants']] and 0 unnamed objects
-> GIVEN_TOPIC_ENTITY: m.03_dwn <- sports.sports_team.team_mascot <- baseball.baseball_league.teams query result: [['National League']] and 0 unnamed objects
-> GIVEN_TOPIC_ENTITY: m.03_dwn -> sports.mascot.team query result: [['San Francisco Giants']] and 0 unnamed objects
\end{lstlisting}

Sample response:

\begin{lstlisting}[basicstyle=\small\ttfamily, breaklines=true]
2014 World Series
\end{lstlisting}

\section{Method Details}
\label{sec:fig}

\begin{figure*}[t]
  \centering 
    \includegraphics[width=\linewidth]{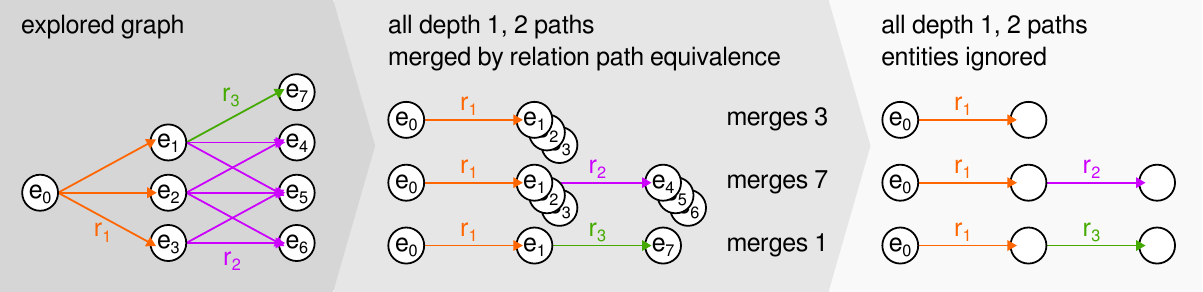}
  \caption{\textbf{Entity-Agnostic Relation Path Pruning} \ The explored subgraph (left) is first grouped by relation path equivalence (center). Then, intermediate entities are ignored, abstracting paths into relational paths (right).}
  \label{fig:relpath}

\end{figure*}

Figure \ref{fig:relpath} illustrates the Entity-Agnostic Relation Path construction.

\section{Overview of Datasets}
\label{sec:dataset}

Table \ref{dataset-table} shows the basic information of datasets used in this thesis.

\begin{table*}[t]
  \centering
  \small
  \begin{tabular}{lllllll}
    \hline
    \textbf{Dataset}          & \textbf{KG} & \textbf{Complexity} & \textbf{Reasoning} & \textbf{Annotation} & \textbf{Source} \\
    \hline
     CWQ &  Freebase  & Complex Qs & Multi-hop & SPARQL, Ans. & Web-derived \\
     WebQSP & Freebase  & Moderate Qs & Single-hop (most) & SPARQL, Ans. & Web-derived \\
     WebQuestions & Freebase  & Simple-Moderate Qs & Single-hop (most) & Ans. only & Web-derived \\
     GrailQA & Freebase  & Complex (varied) & Multi-hop & Logic forms, Ans. & Human-controlled \\
    \hline
  \end{tabular}
  \caption{\label{dataset-table}
    Overview of KGQA Datasets \cite{wang2026kgreasonerreinforcedmodelendtoend}
  }
\end{table*}


\section{Case Study}

Table \ref{tab:case_study_json1} and Table \ref{tab:case_study_json2} are cases that shows the reasoning outcomes of ToG and ToP.

\begin{table*}[h]
\centering

\begin{tabular}{|p{\textwidth}|}
\hline 
Question: "The person who attended education institution in East Whittier Elementary School died where?" \\
\hline
ToG Output\\
\hline 
\small\ttfamily
\begin{verbatim}
{
    "question": "The person who attended education institution in East Whittier Elementary School
                 died where?",
    "results": "{Yes}. Based on the given knowledge triplets, we can identify that Richard Nixon attended
                East Whittier Elementary School. The triplets also state that Richard Nixon died in
                New York City. Therefore, the answer to the question is {New York City}.",
    "reasoning_chains": [
        [
            [
                [
                    "East Whittier Elementary School",
                    "education.educational_institution.students_graduates",
                    "UnName_Entity"
                ]
            ]
        ]
    ]
}
\end{verbatim} \\
\hline
ToP Paths and Output\\
\hline
\small\ttfamily
\begin{verbatim}
Paths:
-> GIVEN_TOPIC_ENTITY: m.0_sp3dx <- education.education.institution -> education.education.student -> 
people.deceased_person.place_of_death query result: [['New York City']] and 
0 unnamed objects
-> GIVEN_TOPIC_ENTITY: m.0_sp3dx <- education.education.institution -> education.education.student <- 
people.place_of_interment.interred_here query result: [['Richard Nixon Library & Birthplace']] and 
0 unnamed objects
-> GIVEN_TOPIC_ENTITY: m.0_sp3dx <- education.education.institution -> education.education.student -> 
people.deceased_person.place_of_burial query result: [['Richard Nixon Library & Birthplace']] and 
0 unnamed objects
-> GIVEN_TOPIC_ENTITY: m.0_sp3dx -> education.educational_institution.students_graduates query result: 
[] and 1 unnamed objects
-> GIVEN_TOPIC_ENTITY: m.0_sp3dx <- education.education.institution query result: [] and 
1 unnamed objects

{"response": "['New York City']"}
\end{verbatim} \\
\hline
\end{tabular}
\caption{ \label{tab:case_study_json1} Both ToP and ToG generated correct answer.}
\end{table*}

\begin{table*}[h]
\centering

\begin{tabular}{|p{\textwidth}|}
\hline 
Question: "Lou Seal is the mascot for the team that last won the World Series when?" \\
\hline
ToG Output\\
\hline 
\small\ttfamily
\begin{verbatim}
{
    "question": "Lou Seal is the mascot for the team that last won the World Series when?",
    "results": "Based on the given knowledge triplets, the mascot for the San Francisco Giants is
                Lou Seal. However, the triplets do not contain information about when the San Francisco Giants
                last won the World Series. Therefore, additional knowledge is required to answer the question.",
    "reasoning_chains": [
        [
            [
                [
                    "Lou Seal",
                    "sports.mascot.team",
                    "San Francisco Giants"
                ]
            ]
        ]
    ]
}
\end{verbatim} \\
\hline
ToP Paths and Output\\
\hline
\small\ttfamily
\begin{verbatim}
Paths:
-> GIVEN_TOPIC_ENTITY: m.03_dwn <- sports.sports_team.team_mascot <- 
sports.sports_championship_event.champion query result: [['2010 World Series'], ['2012 World Series'], 
['2014 World Series']] and 0 unnamed objects
-> GIVEN_TOPIC_ENTITY: m.03_dwn <- sports.sports_team.team_mascot <- 
sports.sports_league_participation.team query result: [] and 3 unnamed objects
-> GIVEN_TOPIC_ENTITY: m.03_dwn <- sports.sports_team.team_mascot query result: 
[['San Francisco Giants']] and 0 unnamed objects
-> GIVEN_TOPIC_ENTITY: m.03_dwn <- sports.sports_team.team_mascot <- baseball.baseball_league.teams 
query result: [['National League']] and 0 unnamed objects
-> GIVEN_TOPIC_ENTITY: m.03_dwn -> sports.mascot.team query result: [['San Francisco Giants']] and 
0 unnamed objects


{"response": "2014 World Series"}
\end{verbatim} \\
\hline
\end{tabular}
\caption{ \label{tab:case_study_json2} ToG failed to generate enough reasoning chains and fail to answer the question. In contrast, ToP explored plenty of paths, which are sufficient to give an answer. }
\end{table*}

\end{document}